\documentclass[twoside,11pt]{article}

\usepackage{paralist,amsmath, amssymb}
\usepackage{booktabs}
\usepackage{multirow}
\usepackage{algorithm,algorithmic}
\usepackage{jmlr2e}

\usepackage[colorlinks,
            linkcolor=blue,
            citecolor=blue,
            urlcolor=magenta,
            linktocpage,
            plainpages=false]{hyperref}

\makeatletter
\newcounter{ALC@tempcntr}
\makeatother

\makeatletter
\AtBeginDocument{\Hy@breaklinkstrue}
\makeatother

\newtheorem{thm}{Theorem}

\newtheorem{cor}[thm]{Corollary}
 \newtheorem{ass}{Assumption}

\def \E {\mathrm{E}}
\def \x {\mathbf{x}}

\def \D {\mathbb{D}}

\def \w {\mathbf{w}}
\def \R {\mathbb{R}}

\def \W {\mathcal{W}}
\def \N {\mathcal{N}}

\def \wt {\widetilde{\w}}

\def \P {\mathbb{P}}

\def \wb {\bar{\w}}

\def \wh {\widehat{\w}}

\def \O {\widetilde{O}}
\def \OMG {\widetilde{\Omega}}

\DeclareMathOperator*{\ess}{ess}
\DeclareMathOperator*{\argmin}{argmin}

\DeclareMathOperator*{\pr}{prior}

\begin{document}

\title{Stochastic Approximation of Smooth and Strongly Convex Functions: Beyond the $O(1/T)$ Convergence Rate}

\author{\name Lijun Zhang \email zhanglj@lamda.nju.edu.cn\\
\name Zhi-Hua Zhou \email zhouzh@lamda.nju.edu.cn\\
       \addr National Key Laboratory for Novel Software Technology\\
       Nanjing University, Nanjing 210023, China
}
\editor{}

\maketitle

\begin{abstract}%
Stochastic approximation (SA) is a classical approach for stochastic convex optimization. Previous studies have demonstrated that the convergence rate of SA can be improved by introducing either smoothness or strong convexity condition.  In this paper, we make use of smoothness and strong convexity  \emph{simultaneously} to boost the convergence rate. Let $\lambda$ be the modulus of strong convexity,  $\kappa$ be the condition number, $F_*$ be the minimal risk, and $\alpha>1$ be some \emph{small} constant. First, we demonstrate that, in expectation, an  $O(1/[\lambda T^\alpha] +  \kappa F_*/T)$ risk bound is attainable when $T =  \Omega(\kappa^\alpha)$. Thus, when $F_*$ is small, the convergence rate could be faster than $O(1/[\lambda T])$ and approaches $O(1/[\lambda T^\alpha])$ in the ideal case. Second, to further benefit from small risk, we show that, in expectation,  an  $O(1/2^{T/\kappa}+F_*)$ risk bound is achievable. Thus, the excess risk reduces exponentially until reaching $O(F_*)$, and if $F_*=0$, we obtain a global linear convergence. Finally, we emphasize that our proof is constructive and each risk bound is equipped with an efficient stochastic algorithm attaining that bound.
\end{abstract}


\section{Introduction}
Stochastic optimization (SO) is  frequently encountered in a vast number of areas, including telecommunication, medicine, and finance, to name but a few \citep{Shapiro:2014:LSP}. SO aims to minimize an objective function which is given in a form of the expectation.  Formally, the problem can be formulated as
\begin{equation} \label{eqn:so}
\min_{\w \in \W} \  F(\w)=\E_{f\sim\P}\left[f(\w)\right]
\end{equation}
where $f(\cdot): \W \mapsto \R$ is a random function sampled from a distribution $\P$. A well-known special case is the risk minimization in machine learning, whose objective function is
\[
F(\w)=\E_{(\x,y) \sim \D} \left[ \ell\big(y, \langle \w, \x\rangle \big) \right]
\]
where $(\x,y)$ denotes  a random instance-label pair sampled from certain distribution $\D$, $\w$ is the model for prediction, and $\ell(\cdot,\cdot)$ is a  loss that measures the prediction error \citep{vapnik-1998-statistical}.

In this paper, we focus on stochastic convex optimization (SCO), in which both the domain $\W$ and the expected function $F(\cdot)$ are convex. A basic difficulty of solving stochastic optimization problem is that the distribution $\P$ is generally unknown, or even if known, it is hard to evaluate the expectation exactly \citep{nemirovski-2008-robust}. To address this challenge,  two different ways have been proposed: sample average approximation (SAA)  \citep{Kim2015} and stochastic approximation (SA) \citep{SA:Springer}. SAA collects a set of random functions $f_1,\ldots,f_T$ from $\P$, and constructs the empirical average $\sum_{i=1}^T f_i(\cdot)/T$ to approximate the expected function $F(\cdot)$. In contrast, SA tackles the stochastic optimization problem directly, at each iteration using a noisy observation of $F(\cdot)$ to improve the current iterate.

Compared with SAA, SA is more efficient due to the low computational cost per iteration, and has received significant research interests from optimization and machine learning communities \citep{Zhang_SVM,JMLR:Adaptive,pmlr-v40-Ge15,pmlr-v65-wang17a}. The performance of SA algorithms is typically measured by the excess risk:
\[
F(\w_T)-\min_{\w \in \W} F(\w)
\]
where  $\w_T$ is the solution returned after $T$ iterations. For Lipschitz continuous convex functions, stochastic gradient descent (SGD) achieves the unimprovable $O(1/\sqrt{T})$ rate of convergence. Alternatively, if the optimization problem has certain curvature properties, then faster rates are sometimes possible. Specifically, for smooth functions, SGD is equipped with an $O(1/T+\sqrt{F_*/T})$ risk bound, where $F_*=\min_{\w \in \W} F(\w)$ is the minimal risk \citep{Smooth:Risk}. Thus, the convergence rate for smooth functions could  be faster than $O(1/\sqrt{T})$ when the minimal risk is small. For strongly convex functions, the convergence rate can also be improved to $O(1/[\lambda T])$, where $\lambda$ is the modulus of strong convexity \citep{COLT:Hazan:2011}.

From the above discussions, we observe that either smoothness or strong convexity could be exploited to improve the convergence rate of SA. This observation motivates subsequent studies that boost the convergence rate by considering  smoothness and strong convexity  \emph{simultaneously}. However, existing results are unsatisfactory because they either rely on strong assumptions \citep{pmlr-v30-Mahdavi13,SGD:Growth}, are only applicable to unconstrained domains \citep{NIPS2011_4316,NIPS2014_5355}, or limited to the problem of finite sum \citep{NIPS2012_SGM,SDCA_Shalev,NIPS13:ASGD}. This paper demonstrates that for the general SO problem, the convergence rate of SA could be faster than $O(1/T)$ when both smoothness and strong convexity are present and the minimal risk is small. Our work is similar in spirit to a recent study of SAA \citep{ERM:COLT:17}, which also establishes faster rates under similar conditions. The main contributions of our paper are summarized below.
\begin{compactitem}
  \item First, we propose a fast algorithm for stochastic approximation (FASA), which applies epoch gradient descent (Epoch-GD) \citep{COLT:Hazan:2011} with carefully designed initial solution and step size. Let $\kappa$ be the condition number and $\alpha>1$ be some small constant. Our theoretical analysis shows that, in expectation, FASA achieves an $O\left(1/[\lambda T^\alpha] +  \kappa  F_*/T\right)$ risk bound when the number of iterations $T =  \Omega(\kappa^\alpha)$.  As a result, the convergence rate could be faster than $O(1/[\lambda T])$ when $F_*$ is small, and approaches $O(1/[\lambda T^\alpha])$ when $F_*=O(1/T^{\alpha-1})$.
  \item Second, to further benefit from small risk, we propose to use a fixed step size in  Epoch-GD, and establish an   $O(1/2^{T/\kappa}+F_*)$ risk bound which holds in expectation. Thus, the excess risk reduces exponentially until reaching $O(F_*)$, and if $F_*=0$, we obtain a global linear convergence.
\end{compactitem}
\section{Related Work}
In this section, we  review related work on SA and SAA.
\subsection{Stochastic Approximation (SA)}
For brevity, we only discuss first-order methods of SA, and results of zero-order methods can be found in the literature \citep{Nesterov_Grad_Free,NIPS:Finite}.

For Lipschitz continuous convex functions, stochastic gradient descent (SGD) exhibits the optimal $O(1/\sqrt{T})$ risk bound \citep{Problem_complexity,zinkevich-2003-online}. When the random function $f(\cdot)$ is nonnegative and smooth, SGD (with a suitable step size) has a risk bound of $O(1/T+\sqrt{F_*/T})$, becoming $O(1/T)$  if the minimal risk $F_*=O(1/T)$ \citep[Corollary 4]{Smooth:Risk}. If the expected function $F(\cdot)$ is $\lambda$-strongly convex, some variants of SGD \citep{COLT:Hazan:2011,JMLR:v15:hazan14a,ICML2012Rakhlin,ICML2013Shamir:ICML} achieve an $O(1/[\lambda T])$ rate which is known to be minimax optimal~\citep{IT:SCO}. For the square loss and the logistic loss, an $O(1/T)$ rate is attainable without strong convexity \citep{NIPS2013_4900}. When the random function $f(\cdot)$ is $\eta$-exponentially concave,  the online Newton step (ONS) is equipped with an $\O(d/ [\eta T])$ risk bound, where $d$ is the dimensionality \citep{ML:Hazan:2007,COLT:2015:Mahdavi}.
When the expected function is both smooth and strongly convex, we still have the $O(1/T)$ convergence rate but with a smaller constant  \citep{Lan:SCSC}. Specifically, the constant in the big O notation depends on the \emph{variance} of the stochastic gradient instead of the maximum norm.

There are some studies that have established convergence rates that are faster than $O(1/T)$ when both smoothness and strong convexity are present.
\cite{NIPS2011_4316} and \cite{NIPS2014_5355} demonstrate that the distance between the SGD iterate and the optimal solution decreases at a linear rate in the beginning, but their results are limited to unconstrained problems. When an upper bound of $F_*$ is available,  \cite{pmlr-v30-Mahdavi13} show that it is possible to reduce the excess risk at a linear rate until certain level. Under a strong growth condition, \cite{SGD:Growth} prove that SGD could achieve a global linear rate. Recently, a variety of variance reduction techniques have been proposed and yield faster rates for SA \citep{NIPS2012_SGM,SDCA_Shalev,NIPS13:ASGD}. However, these methods are restricted to the special case that the expected function is a finite sum, and thus cannot be applied if the distribution is unknown. As can be seen, existing fast rates of SA are restricted to special problems or rely on strong assumptions. We will provide detailed comparisons in  Section \ref{Sec:Main} to illustrate the advantage of this study---our setting is more general and our convergence rates are faster.

While our paper focuses on stochastic convex optimization, we note there has been a recent surge of interests in developing SA algorithms for non-convex problems  \citep{pmlr-v40-Ge15,pmlr-v48-allen-zhua16,SVRG:Nonconvex,pmlr-v65-zhang17b}.
\subsection{Sample Average Approximation (SAA)}
SAA is also referred to as empirical risk minimization (ERM) in  machine learning. In the literature, there are plenty of theories for SAA \citep{Kim2015} or ERM \citep{vapnik-1998-statistical}. In the following, we only discuss related work on SAA in the past decade.

To present the results in SAA, we use $T$ to denote the total number of training samples. When the random function $f(\cdot)$ is Lipschitz continuous, \cite{COLT:Shalev:2009} establish an $\O(\sqrt{d/T})$ risk bound. When $f(\cdot)$ is $\lambda$-strongly convex and Lipschitz continuous,  \cite{COLT:Shalev:2009} further prove an $O(1/[\lambda T])$ risk bound which holds in expectation. When $f(\cdot)$ is $\eta$-exponentially concave,  an $\O(d/[\eta T])$ risk bound is attainable \citep{NIPS2015_Exp,arXiv:1605.01288}. Lower bounds of ERM for stochastic optimization are investigated by \cite{NIPS2016_ERM}. In a recent work, \cite{ERM:COLT:17} establish an $\O(d/T + \sqrt{F_*/T})$ risk bound when $f(\cdot)$ is smooth and $F(\cdot)$ is Lipschitz continuous. The most surprising result is that when $f(\cdot)$ is smooth and $F(\cdot)$ is Lipschitz continuous and $\lambda$-strongly convex,  \cite{ERM:COLT:17} prove an $O(1/[\lambda T^2] + \kappa F_*/T)$ risk bound,  when $T=\OMG(\kappa d)$. Thus, the convergence rate of ERM could be faster than $O(1/[\lambda T])$ when both smoothness and strong convexity are present and the number of training samples is large enough.

\section{Our Results} \label{Sec:Main}
We first introduce assumptions used in our analysis, then present our algorithms and theoretical guarantees.

\subsection{Assumptions}
\begin{ass}\label{ass:0}
The random function $f(\cdot)$ is nonnegative.
\end{ass}

\begin{ass}\label{ass:1}
The random function $f(\cdot)$ is (almost surely) $L$-smooth over $\W$, that is,
\begin{equation} \label{eqn:f:smooth}
\left \|\nabla f(\w)-\nabla f(\w') \right\| \leq L \|\w-\w'\|, \ \forall \w, \w' \in \W.
\end{equation}
\end{ass}

\begin{ass}\label{ass:2}
The expected function $F(\cdot)$ is $\lambda$-strongly convex over $\W$, that is,
\begin{equation} \label{eqn:F:strong}
   F(\w)+\langle \nabla F(\w), \w' -\w \rangle + \frac{\lambda}{2} \|\w'-\w\|^2 \leq F(\w'), \ \forall \w, \w' \in \W.
\end{equation}
\end{ass}

\begin{ass}\label{ass:3}
The gradient of the random function is (almost surely) upper bounded by $G$, that is,
\begin{equation} \label{eqn:bounded:grad}
\| \nabla f(\w)\| \leq G, \ \forall \w \in \W.
\end{equation}
\end{ass}

\paragraph{Remark 1} We have the following comments regarding our assumptions.
\begin{compactitem}
\item The above assumptions hold for many popular machine learning problems, such as  (regularized) linear regression or logistic regression.
\item Based on Assumptions~\ref{ass:1} and \ref{ass:2}, we define the condition number $\kappa=L/\lambda$, which will be used to characterize the performance of our methods. For simplicity, we assume $L$ is a constant, and thus $\kappa$ and $1/\lambda$ are on the same order.
\item Let $\w_* =\argmin_{\w \in \W} F(\w)$ be the optimal solution to (\ref{eqn:so}).  Assumption~\ref{ass:2} implies  \citep{COLT:Hazan:2011}
\begin{equation} \label{eqn:stron:con}
\frac{\lambda}{2} \|\w - \w_*\|^2 \leq F(\w)-F(\w_*), \ \forall \w \in \W.
\end{equation}
Actually, in our analysis, we only make use of (\ref{eqn:stron:con}) instead of (\ref{eqn:F:strong}).
\end{compactitem}

\begin{algorithm}[t]
\caption{Epoch Gradient Descent (Epoch-GD)}
{\bf Input}: parameters $\eta_1$, $T_1$, $T$, and $\w_0$
\begin{algorithmic}[1]
\STATE Initialize $\w_1^1 =\w_0$, and set $k=1$
\WHILE{$\sum_{i=1}^k T_i \leq T$}
\FOR{$t=1$ to $T_k$}
\STATE Sample a random function $f_t^k(\cdot)$ from $\P$
\STATE Update
\[
\w_{t+1}^k=\Pi_{\W}\left[\w_t^k-\eta_k \nabla f_t^k(\w_t^k) \right]
\]
\ENDFOR
\STATE $\w_1^{k+1}=\frac{1}{T_k} \sum_{t=1}^{T_k} \w_t^k$
\STATE $T_{k+1}=2T_k$ and $\eta_{k+1}=\eta_k/2$
\STATE $k=k+1$
\ENDWHILE
\RETURN $\w_1^{k}$
\end{algorithmic}\label{alg:1}
\end{algorithm}
\subsection{A General Algorithm}
We first introduce a general algorithm for SA, which always achieves an $O(1/\lambda T)$ rate, and becomes faster when $F_*$ is small.

\subsubsection{Fast Algorithm for Stochastic Approximation (FASA)}
Our fast algorithm for stochastic approximation (FASA) takes epoch gradient descent (Epoch-GD) as a subroutine.  Although \cite{COLT:Hazan:2011} have established the convergence rate of Epoch-GD under the strong convexity condition, they did not utilize smoothness in their analysis. The procedures of  Epoch-GD and FASA are described in Algorithm~\ref{alg:1} and Algorithm~\ref{alg:2}, respectively.

Epoch-GD is an extension of stochastic gradient descent (SGD). It divides the optimization process into a sequence of epochs. In each epoch, Epoch-GD applies SGD multiple times, and the averaged iterate is passed to the next epoch. In the algorithm, we use
 $\Pi_{\W}[\cdot]$ to denote the projection onto the nearest point in $\W$. There are $4$ input parameters of Epoch-GD: (1) $\eta_1$, the step size used in the first epoch; (2) $T_1$, the size of the first epoch;  (3) $T$, the total number of stochastic gradients that can be consumed; and (4) $\w_0$, the initial solution. In each consecutive epoch, the step size decreases exponentially and the size of epoch increases exponentially.

\begin{algorithm}[t]
\caption{Fast Algorithm for Stochastic Approximation (FASA)}
{\bf Input}: parameters $L$, $\lambda$, $T$, and $\alpha$
\begin{algorithmic}[1]
\STATE Let $\wb$ be any point in $\W$, and set $\kappa=L/\lambda$
\STATE Invoke Epoch-GD($1/\lambda$,$4$,$T/2$, $\wb$), and denote the solution by $\wh$
\STATE Invoke Epoch-GD($1/4 L$,$2^{\alpha+3} \kappa$,$T/2$, $\wh$), and denote the solution by $\wt$
\RETURN $\wt$
\end{algorithmic}\label{alg:2}
\end{algorithm}
In FASA, we first invoke Epoch-GD with an arbitrary initial solution, and the number of stochastic gradients is set to be $T/2$. The purpose of this step is to get a good solution $\wh$ at the expense of $T/2$ stochastic gradients.\footnote{In this step, Epoch-GD can be replaced with any  algorithm that achieves the optimal $O(1/\lambda T)$ rate for strongly convex stochastic optimization, e.g., SGD with $\alpha$-suffix averaging \citep{ICML2012Rakhlin}.} Then, Epoch-GD is invoked again with $\wh$ as its initial solution and a budget of $T/2$ stochastic gradients. This time, we set a large epoch size to utilize the fact that the initial solution is of high quality. The convergence rate of FASA is given below.

\begin{thm} \label{thm:1} Suppose
\begin{equation} \label{eqn:lower:T}
T \geq \kappa^\alpha
\end{equation}
where $\alpha > 1$ is some constant. Under Assumptions~\ref{ass:0}, \ref{ass:1}, \ref{ass:2} and \ref{ass:3}, the solution $\wt$ returned by Algorithm~\ref{alg:2} satisfies
\[
\E\left[F(\wt)\right] - F_* \leq  \frac{2^{\alpha^2+5\alpha+5} G^2}{\lambda T^\alpha} + \frac{2^{2 \alpha+5} \kappa  F_*}{(2^{\alpha-1}-1)T}
\]
where $F_*=F(\w_*)=\min_{\w \in \W} F(\w)$ is the minimal risk.
\end{thm}
\paragraph{Remark 2} The above theorem implies that when $T$ is large enough, i.e., $T =  \Omega(\kappa^\alpha)$, FASA achieves an
 \[
 O\left(  \frac{1}{\lambda T^\alpha} + \frac{ \kappa  F_*}{T}\right)
 \]
rate of convergence, which is faster than $O(1/[\lambda T])$ when the minimal risk is small. In particular, when $F_*=O(1/T^{\alpha-1})$, the convergence rate is improved to  $O(1/[\lambda T^\alpha])$. Note that the upper bound has an exponential dependence on $\alpha$, so it is meaningful only when $\alpha$ is chosen as a \emph{small} constant.

\paragraph{Remark 3} Note that our algorithm is translation-invariant, i.e., it  does not change if we translate the function by a constant. Since the upper bound in Theorem~\ref{thm:1} depends on the minimal risk $F_*$, one may attempt to subtract a constant from the function to make the bound tighter. However, because of the nonnegative requirement in Assumption~\ref{ass:0}, the best we can do is to redefine
\[
f(\w) \leftarrow f(\w)- \ess \inf_{f \sim \P} \inf_{\w \in \W} f(\w)
\]
and replace $F_*$ in Theorem~\ref{thm:1} with $F_*- \ess \inf_{f \sim \P} \inf_{\w \in \W} f(\w)$.

To simplify Theorem~\ref{thm:1}, we provide the following corollary by setting $\alpha=2$.
\begin{cor} \label{cor:1}
Suppose $T \geq \kappa^2$. Under the same conditions as Theorem~\ref{thm:1}, we have
\[
\E\left[F(\wt)\right] - F_* \leq  \frac{2^{19} G^2}{\lambda T^2} +\frac{2^{9} \kappa  F(\w_*)}{T}=O\left(\frac{1}{\lambda T^2} +\frac{\kappa  F_*}{T} \right).
\]
\end{cor}

\subsubsection{Comparisons with Previous Results} \label{sec:com:res}
In the following, we compare our Theorem~\ref{thm:1} and Corollary~\ref{cor:1} with related work in SA \citep{Lan:SCSC,NIPS2011_4316,NIPS2014_5355} and SAA \citep{ERM:COLT:17}.

For smooth and strongly convex functions, \citet[Proposition 9]{Lan:SCSC} have established an $O(1/T^2+\sigma^2/[\lambda T])$ rate for the expected risk, where $\sigma^2$ is the variance of the stochastic gradient. Note that this rate is worse than that in  Corollary~\ref{cor:1} because $\sigma^2$ is a constant in general, even when $F_*$ is small. For example, consider the problem of linear regression
\[
\min_{\w \in \W} \  F(\w)=\E_{(\x,y) \sim \D} \left[ (\x^\top \w-y)^2 \right],
\]
and assume $y=\x^\top \w_*+ \epsilon$ where $\epsilon \sim \N(0, \rho^2)$ is the Gaussian random noise and $\w_*\in \W$. Then, $F_*=\E[\epsilon^2]=\rho^2$, which approaches zero as $\rho \rightarrow 0$. On the other hand, the variance of the stochastic gradient at solution $\w_t$ can be decomposed as
\[
\begin{split}
\sigma^2=&\E\left[\left\| 2(\x^\top \w_t-y) \x - \E\big[2(\x^\top \w_t-y) \x \big] \right\|^2 \right] \\
=& 4 \E \left[\left\|\big(\x \x^\top -\E[\x \x^\top] \big) (\w_t- \w_*)\right\|^2 \right]+ 4\E \left[ \|\epsilon \x\|^2 \right].
\end{split}
\]
Even there is no noise, i.e., $\rho=0$, the variance is nonzero due to the randomness of $\x$.

For unconstrained problems, \citet{NIPS2011_4316} and \citet{NIPS2014_5355} have analyzed the distance between the SGD iterate and the optimal solution under the smoothness and strong convexity condition. In particular, Theorem 1 of \citet{NIPS2011_4316} (with $\alpha=1$ and $\mu C=2$) implies the following convergence rate for the expected risk
\[
O\left( \frac{\exp(\kappa^2)}{n^2} + \frac{F_* \log T}{\lambda^2 T}\right)
\]
which is worse than our Corollary~\ref{cor:1} because of the additional $\log T/\lambda$ factor in the second term. Theorem 2.1 of \citet{NIPS2014_5355} leads to the following rate
\begin{equation}\label{eqn:rate:nips:2014}
O\left( \left(1-\frac{\lambda}{T}\right)^T +\frac{\kappa F_*}{T}\right)
\end{equation}
which is also worse than our Corollary~\ref{cor:1} because $(1-\lambda/T)^T$ becomes a constant when $T\rightarrow \infty$. We note that it is possible to extend the analysis of \citet{NIPS2014_5355} to constrained problems, but the convergence rate becomes slower, and thus is worse than our rate. Detailed discussions about how to simplify and extend the result of \citet{NIPS2014_5355} are provided in Appendix \ref{sec:SGD}.

The convergence rate in Corollary~\ref{cor:1} matches the state-of-the-art convergence rate of SAA \citep{ERM:COLT:17}. Specifically, under similar conditions,  \citet[Theorem 3]{ERM:COLT:17} have proved an $O(1/[\lambda T^2] + \kappa F_*/T)$ risk bound for SAA, when $T=\OMG(\kappa d)$. Compared with the results of \citet{ERM:COLT:17}, our theoretical guarantees have the following advantages:
\begin{compactitem}
  \item The lower bound of $T$ in our results is independent from the dimensionality, and thus our results can be applied to infinite dimensional problems, e.g., learning with kernels. In contrast, the lower bound of $T$ given by  \citet[Theorem 3]{ERM:COLT:17} depends on the dimensionality.
  \item For the special problem of supervised learning,  \citet[Theorem 7]{ERM:COLT:17} shows that the lower bound on $T$ can be replaced with $\Omega(\kappa^2)$. However, it does not support the case $T\in (\kappa, \kappa^2)$, which is covered by our Theorem~\ref{thm:1}.
  \item The convergence rate in Theorem~\ref{thm:1} keeps improving as $\alpha$ increases. As a result, when $\alpha>2$, the convergence rate in Theorem~\ref{thm:1} is faster than that of SAA given by \cite{ERM:COLT:17}.
\end{compactitem}

\subsection{A Special Algorithm for Small Risk}
The convergence rate of FASA cannot go beyond $O(1/[\lambda T^\alpha])$, even when $F_*$ is $0$.  In the following, we develop a special algorithm for the case that $F_*$ is small. The new algorithm achieves a linear convergence when $F_*$ is small, although it may not perform well otherwise.

\subsubsection{Epoch Gradient Descent with Fixed Step Size (Epoch-GD-F)}
The new algorithm is a variant of Epoch-GD, in which the step size, as well as the size of each epoch, is fixed. We name the new algorithm as epoch gradient descent with fixed step size (Epoch-GD-F), and summarize it in Algorithm~\ref{alg:3}. Epoch-GD-F has $4$ parameters: (1) $\eta$, the fixed step size; (2) $T'$, the size of each epoch; (3) $T$, the total number of stochastic gradients that can be consumed; and (4) $\w_0$, the initial solution. We bound the excess risk of Epoch-GD-F in the following theorem.

\begin{algorithm}[t]
\caption{Epoch Gradient Descent with Fixed Step Size (Epoch-GD-F)}
{\bf Input}: parameters $\eta$, $T'$, $T$, and $\w_0$
\begin{algorithmic}[1]
\STATE Set $\w_1^1 =\w_0$  and $k=1$
\WHILE{$k \leq T/T'$}
\FOR{$t=1$ to $T'$}
\STATE Sample a random function $f_t^k(\cdot)$ from $\P$
\STATE Update
\[
\w_{t+1}^k=\Pi_{\W}\left[\w_t^k-\eta \nabla f_t^k(\w_t^k) \right]
\]
\ENDFOR
\STATE $\w_1^{k+1}=\frac{1}{T'} \sum_{t=1}^{T'} \w_t^k$
\STATE $k=k+1$
\ENDWHILE
\RETURN $\wt=\w_1^{k}$
\end{algorithmic}\label{alg:3}
\end{algorithm}
\begin{thm} \label{thm:2} Set
\begin{equation} \label{eqn:para:second}
\eta= \frac{1}{4 \beta L}, \ T'= 16 \beta \kappa
\end{equation}
where $\beta>1$ is some constant, and $\w_0$ be any point in $\W$. Under Assumptions~\ref{ass:0}, \ref{ass:1} and \ref{ass:2}, the solution $\wt$ returned by Algorithm~\ref{alg:3} satisfies
\[
\E\left[F(\wt)\right] - F_* \leq  \frac{F(\w_0)-F_*}{2^{k^\dag}}+ \frac{2  F_*}{\beta}
\]
where $k^\dag = \lfloor T/ T' \rfloor$.
\end{thm}
\paragraph{Remark 4} From the above theorem, we observe that the excess risk is upper bounded by two terms: the first one decreases \emph{exponentially} w.r.t.~the number of epoches and the second one depends on $F_*$. When $\beta=O(1)$, the excess risk is on the order of
\[
O\left( \frac{1}{2^{T/\kappa}}+  F_* \right)
\]
which means it reduces exponentially until reaching $O(F_*)$. Note that if $F_*=0$, we obtain a global linear convergence.

To better illustrate the convergence rate in Theorem~\ref{thm:2}, we present the iteration complexity of Epoch-GD-F.
\begin{cor} \label{cor:small} Assume
\[
T=\Omega\left(\beta \kappa \log \frac{1}{\epsilon} \right).
\]
Under the same condition as Theorem~\ref{thm:2}, the solution $\wt$ returned by Algorithm~\ref{alg:3} satisfies
\[
\E\left[F(\wt)\right] - F_* \leq  \epsilon + \frac{2 F_*}{\beta}.
\]
\end{cor}
\subsubsection{Comparisons with Previous Results}
In the following, we compare our Theorem~\ref{thm:2} and Corollary~\ref{cor:small}  with related work in SA \citep{pmlr-v30-Mahdavi13,SGD:Growth,NIPS2011_4316,NIPS2014_5355}.

When a prior knowledge $\epsilon_{\pr} \geq F_*$ is given beforehand,  \cite{pmlr-v30-Mahdavi13} show that when
\[
T =\Omega\left( d \beta^3 \kappa^4 \log \frac{1}{\epsilon} \right),
\]
their stochastic algorithm is able to find a solution $\wh$ such that with high probability
\[
F(\wh) \leq \epsilon_{\pr} +  \epsilon +   \frac{2 \epsilon_{\pr}}{\beta}.
\]
Although our Corollary~\ref{cor:small} only holds in expectation, it is stronger than that of \cite{pmlr-v30-Mahdavi13}  in the following aspects:
\begin{compactitem}
\item Their algorithm needs a prior knowledge $\epsilon_{\pr} \geq F_*$, while our algorithm does not.
\item The final risk of their solution is upper bounded in terms of $\epsilon_{\pr}$, while in our case, the risk is upper bounded in terms of  $F_*$, which is smaller than $\epsilon_{\pr}$.
\item Their sample complexity has a linear dependent on the dimensionality $d$, in contrast ours is dimensionality-independent. Thus, our results can be applied to the non-parametric setting where hypotheses lie in a functional space of infinite dimension.
\item The dependence of their sample complexity on $\beta$ and $\kappa$ is much higher than ours.
\end{compactitem}

Under a strong growth condition \citep{Solodov1998}, \cite{SGD:Growth} have established the following linear convergence rate for SGD when applied to unconstrained problems:
\[
O\left( \left(1-\frac{1}{\kappa}\right)^T \right).
\]
This strong growth condition requires that all stochastic gradients are $0$ at $\w_*$, which is itself a necessary condition for $F_*=0$, because all the random functions are nonnegative. In this case, our Theorem~\ref{thm:2} also achieves a linear rate at the same order. However, our results have the following advantages:
\begin{compactitem}
\item Our Theorem~\ref{thm:2} is more general because it covers the cases that $F_*$ is nonzero.
\item Our results are applicable even when there is a domain constraint.
\end{compactitem}

For unconstrained problems,  Theorem 2.1 of \citet{NIPS2014_5355} with a suitable step size also implies the following rate
\begin{equation}\label{eqn:rate:nips:2014:second}
O\left( \left(1-\frac{1}{\kappa}\right)^T + \kappa F_*\right)
\end{equation}
which is slower than our $O(2^{-T/\kappa} + F_*)$ rate in Theorem~\ref{thm:2}, because of the additional dependence on $\kappa$ in the second term. Besides,  \citet[(2.4) and (2.2)]{NIPS2014_5355} provided the iteration complexity of their algorithm, as well as that of  \citet{NIPS2011_4316} when the minimal risk $F_*$ is \emph{known}. Specifically, the iteration complexities of \citet{NIPS2011_4316} and \citet{NIPS2014_5355} for finding an $\epsilon$-optimal solution are
\begin{equation} \label{eqn:iter:com}
\Omega\left( \log \frac{1}{\epsilon} \left( \kappa^2 + \frac{\kappa^2 F_*}{\epsilon}\right) \right) \textrm{ and } \Omega\left( \log \frac{1}{\epsilon} \left( \kappa + \frac{\kappa^2 F_*}{ \epsilon}\right) \right),
\end{equation}
respectively. In this case, our Theorem~\ref{thm:2} with $\beta=\max(1, 4 F_*/\epsilon)$ implies the following iteration complexity
\begin{equation} \label{eqn:iter:our:fast}
\Omega \left(\log \frac{1}{\epsilon}  \left( \kappa + \frac{ \kappa F_*}{\epsilon} \right)  \right).
\end{equation}
Compared with the lower bounds in (\ref{eqn:iter:com}), our iteration complexity is better because (i) it has a smaller dependence on $\kappa$, and (ii) it  holds for constrained problems.

\section{Analysis}
Our analysis follows from well-known and standard techniques, including the analysis of stochastic gradient descent \citep{zinkevich-2003-online}, self-bounding property of smooth functions \citep{Smooth:Risk}, and the implication of strong convexity \citep{COLT:Hazan:2011}.
\subsection{Proof of Theorem~\ref{thm:1}}
We first state the excess risk of $\wh$, the solution returned by the first call of Epoch-GD. From Theorem 5 of \cite{JMLR:v15:hazan14a}, we have
\begin{equation} \label{eqn:wh}
\E \left[F(\wh) \right] - F(\w_*) \leq \frac{32 G^2}{\lambda T} \overset{\text{(\ref{eqn:lower:T})}}{\leq} \frac{32 G^2}{\lambda \kappa^\alpha}.
\end{equation}

We proceed to analyze the solution returned by the second call of Epoch-GD. In each epoch, the standard stochastic gradient descent (SGD) \citep{zinkevich-2003-online} is applied. The following lemma shows how the excess risk decreases in each epoch.
\begin{lemma} \label{lem:inner}
Apply $T$ iterations of the update
\[
\w_{t+1} = \Pi_{\W}\big[\w_t-\eta \nabla f_t(\w_t) \big]
\]
where $f_t(\cdot)$ is a random function sampled from $\P$, and $\eta < 1/(2L)$. Assume $F(\cdot)$ is convex and  Assumptions~\ref{ass:0} and \ref{ass:1} hold, for any $\w \in \W$, we have
\[
\E\left[ F(\wb)\right] - F(\w) \leq \frac{1}{2 \eta T (1-2 \eta L)}  \E \left[\|\w_1-\w\|^2\right] + \frac{2 \eta L }{(1-2 \eta L)} F(\w)
\]
where $\wb=\frac{1}{T} \sum_{t=1}^T \w_t$.
\end{lemma}

Based on the above lemma, we establish the following result for bounding the excess risk of the intermediate iterate.
\begin{lemma} \label{lemma:inter} Consider the second call of Epoch-GD with parameters ($1/4 L$,$2^{\alpha+3} \kappa$,$T/2$, $\wh$). For any $k$, we have
\begin{equation} \label{eqn:risk:epoch}
\E[F(\w_1^{k+1})] -F(\w_*) \leq \frac{2^{\alpha^2+2\alpha+5} G^2}{\lambda (T_{k})^\alpha} + \frac{2^{\alpha+3} \kappa  F(\w_*)}{ T_{k}} \left( \sum_{i=1}^{k}  \frac{1}{2^{(i-1)(\alpha-1)}} \right) .
\end{equation}
\end{lemma}
The number of epochs made is given by the largest value of $k$ satisfying $\sum_{i=1}^k T_i \leq T/2$, i.e.,
\[
\sum_{i=1}^k T_i =T_1 \sum_{i=1}^k  2^{i-1} = T_1 (2^k-1) \leq \frac{T}{2}.
\]
This value is
\[
k^\dag= \left \lfloor  \log_2 \left(\frac{T}{2 T_1} +1\right) \right\rfloor,
\]
and the final solution is $\wt=\w_1^{k^\dag+1}$. From Lemma~\ref{lemma:inter},  we have
\[
\begin{split}
& F(\w_1^{k^\dag+1})-F(\w_*) \\
\leq & \frac{2^{\alpha^2+2\alpha+5} G^2}{\lambda (T_{k^\dag})^\alpha} + \frac{2^{\alpha+3} \kappa  F(\w_*)}{ T_{k^\dag}} \left( \sum_{i=1}^{k^\dag}  \frac{1}{2^{(i-1)(\alpha-1)}} \right) \\
\leq & \frac{2^{\alpha^2+2\alpha+5} G^2}{\lambda (T_{k^\dag})^\alpha} + \frac{2^{\alpha+3} \kappa  F(\w_*)}{ T_{k^\dag}} \frac{2^{\alpha-1}}{2^{\alpha-1}-1} \\
\leq & \frac{2^{\alpha^2+5\alpha+5} G^2}{\lambda T^\alpha} + \frac{2^{2 \alpha+5} \kappa  F(\w_*)}{(2^{\alpha-1}-1)T}
\end{split}
\]
where the last step is due to
\[
T_{k^\dag} = T_1 2^{k^\dag-1} \geq \frac{T_1}{4} \left(\frac{T}{2 T_1} +1 \right) \geq  \frac{T}{8}.
\]

\subsection{Proof of Lemma~\ref{lem:inner}}
We first introduce the self-bounding property of smooth functions \cite[Lemma 4.1]{Smooth:Risk}.
\begin{lemma} \label{lem:smooth} For an $H$-smooth and nonnegative function $f: \W \mapsto \R$,
\[
\| \nabla f(\w)\| \leq \sqrt{4 H f(\w)}, \ \forall \w \in \W.
\]
\end{lemma}
Assumptions~\ref{ass:0} and \ref{ass:1} imply $f_t(\cdot)$ is nonnegative and $L$-smooth. From Lemma~\ref{lem:smooth}, we have
\begin{equation} \label{eqn:smooth:key}
\|\nabla f_i(\w)\|^2 \leq 4 L  f_i (\w), \ \forall \w \in \W.
\end{equation}

Let $\w_{t+1}'=\w_t - \eta \nabla f_t(\w_t)$. Following the analysis of online gradient descent \citep{zinkevich-2003-online}, for any $\w \in \W$, we have
\[
\begin{split}
& F(\w_{t}) - F(\w) \\
\leq & \langle \nabla F(\w_t), \w_t - \w\rangle  \\
= & \langle \nabla f_t(\w_t), \w_t - \w\rangle + \langle \nabla F(\w_t)-\nabla f_t(\w_t), \w_t - \w \rangle \\
= & \frac{1}{2 \eta} \left( \|\w_t-\w\|^2 - \|\w_{t+1}'-\w\|^2 \right) + \frac{\eta}{2 } \|\nabla f_t(\w_t)\|^2  + \langle \nabla F(\w_t)-\nabla f_t(\w_t), \w_t - \w \rangle \\
\leq & \frac{1}{2 \eta} \left( \|\w_t-\w\|^2 - \|\w_{t+1}-\w\|^2 \right) + \frac{\eta}{2 } \|\nabla f_t(\w_t)\|^2  + \langle \nabla F(\w_t)-\nabla f_t(\w_t), \w_t - \w \rangle \\
\overset{\text{(\ref{eqn:smooth:key})}}{\leq} &  \frac{1}{2 \eta} \left( \|\w_t-\w\|^2 - \|\w_{t+1}-\w\|^2 \right) + 2 \eta L   f_t (\w_t)  + \langle \nabla F(\w_t)-\nabla f_t(\w_t), \w_t - \w \rangle
\end{split}
\]
where the second inequality is due to the nonexpanding property of the projection operator \citep[(1.5)]{nemirovski-2008-robust}.

Summing up over all $t=1,\ldots,T$, we get
\[
\begin{split}
&\sum_{t=1}^T \big( F(\w_{t}) - F(\w) \big) \\
\leq &  \frac{1}{2 \eta}  \|\w_1-\w\|^2 + 2 \eta L \sum_{t=1}^T   f_t (\w_t)  + \sum_{t=1}^T  \langle \nabla F(\w_t)-\nabla f_t(\w_t), \w_t - \w \rangle.
\end{split}
\]
Recall that $F(\cdot)=\E [f_t(\cdot)]$ and $\w_t$ is independent from $f_t$. Taking expectation over both sides, we have
\[
\E\left[\sum_{t=1}^T \big( F(\w_{t}) - F(\w) \big)\right] \leq \frac{1}{2 \eta}  \E \left[\|\w_1-\w\|^2\right] + 2 \eta L \E \left[\sum_{t=1}^T   F (\w_t) \right].
\]
Rearranging the above inequality, we obtain
\[
\E\left[\sum_{t=1}^T  \big( F(\w_{t}) - F(\w) \big)\right]
 \leq \frac{1}{2 \eta (1-2 \eta L)}  \E \left[\|\w_1-\w\|^2\right] + \frac{2 \eta L T}{(1-2 \eta L)} F(\w).
\]

Dividing both sides by $T$, we have
\[
\begin{split}
& \frac{1}{2 \eta T (1-2 \eta L)}  \E \left[\|\w_1-\w\|^2\right] + \frac{2 \eta L}{(1-2 \eta L)} F(\w) \\
\geq & \frac{1}{T} \E\left[\sum_{t=1}^T  \big( F(\w_{t}) - F(\w) \big)\right]  \geq \E\left[ F(\wb)\right] - F(\w)
\end{split}
\]
where the last step is due to Jensen's inequality.
\subsection{Proof of Lemma~\ref{lemma:inter}}
Recall that the following parameters are used in the second call of Epoch-GD
\[
\eta_1 = \frac{1}{4 L}, \ T_1 = 2^{\alpha+3} \kappa,  \ T_{k+1}=2 T_k, \ \eta_{k+1}= \frac{\eta_k}{2}, \ k \geq 1 .
\]
Then, we have
\begin{eqnarray}
\eta_k L \leq \eta_1 L  = \frac{1}{4}, \label{eqn:eta:1}\\
\lambda \eta_k T_k  = 2^{\alpha+1} .\label{eqn:eta:2}
\end{eqnarray}

We prove this lemma by induction on $k$. When $k=1$, from Lemma~\ref{lem:inner}, we have
\[
\begin{split}
&\E\left[ F(\w_1^{2})\right] - F(\w_*) \\
\leq &\frac{1}{2 \eta_1 T_1 (1-2 \eta_1 L)}  \E \left[\|\w_{1}^1-\w_*\|^2\right] + \frac{2 \eta_1 L }{(1-2 \eta_1 L)} F(\w_*)\\
\overset{\text{(\ref{eqn:eta:1})}}{=} & \frac{1}{\eta_1 T_1}  \E \left[\|\w_{1}^1-\w_*\|^2\right] + 4 \eta_1 L  F(\w_*) \\
\overset{\text{(\ref{eqn:eta:2})}}{=} & \frac{\lambda}{2^{\alpha+1}}  \E \left[\|\w_{1}^1-\w_*\|^2\right] + \frac{2^{\alpha+3} \kappa  F(\w_*)}{T_1}  \\
\overset{\text{(\ref{eqn:stron:con})}}{\leq} &  \frac{\lambda}{2^{\alpha+1}}  \frac{2}{\lambda} \E \big[F(\w_{1}^1)-F(\w_*) \big] + \frac{2^{\alpha+3} \kappa  F(\w_*)}{ T_{1}}  \\
\overset{\text{(\ref{eqn:wh})}}{\leq} &  \frac{1}{2^{\alpha}}  \left( \frac{32 G^2}{\lambda \kappa^\alpha} \right) + \frac{2^{\alpha+3} \kappa  F(\w_*)}{ T_{1}}  \\
\overset{(T_1 = 2^{\alpha+3} \kappa)}{=} &   \frac{2^{\alpha^2+2\alpha+5} G^2}{\lambda (T_{1})^\alpha} +  \frac{2^{\alpha+3} \kappa F(\w_*)}{ T_{1}}.
\end{split}
\]

Assume that (\ref{eqn:risk:epoch}) is true for some $k \geq 1$, and we prove the inequality for $k+1$. According to  Lemma~\ref{lem:inner}, we have
\[
\begin{split}
&\E\left[ F(\w_1^{k+2})\right] - F(\w_*) \\
\leq &\frac{1}{2 \eta_{k+1} T_{k+1} (1-2 \eta_{k+1} L)}  \E \left[\|\w_{1}^{k+1}-\w_*\|^2\right] + \frac{2 \eta_{k+1} L }{(1-2 \eta_{k+1} L)} F(\w_*)\\
\overset{\text{(\ref{eqn:eta:1})}}{\leq} & \frac{1}{\eta_{k+1} T_{k+1}}  \E \left[\|\w_{1}^{k+1}-\w_*\|^2\right] + 4 \eta_{k+1} L  F(\w_*) \\
\overset{\text{(\ref{eqn:eta:2})}}{=} & \frac{\lambda}{2^{\alpha+1}}  \E \left[\|\w_{1}^{k+1}-\w_*\|^2\right] + \frac{2^{\alpha+3} \kappa  F(\w_*)}{ T_{k+1}}  \\
\overset{\text{(\ref{eqn:stron:con})}}{\leq} &  \frac{\lambda}{2^{\alpha+1}}  \frac{2}{\lambda} \E \left[ F(\w_{1}^{k+1})-F(\w_*)\right] + \frac{2^{\alpha+3} \kappa  F(\w_*)}{ T_{k+1}}  \\
\overset{\text{(\ref{eqn:risk:epoch})}}{\leq} &  \frac{1}{2^{\alpha}}   \left( \frac{2^{\alpha^2+2\alpha+5} G^2}{\lambda (T_{k})^\alpha} + \frac{2^{\alpha+3} \kappa  F(\w_*)}{ T_{k}} \left( \sum_{i=1}^{k}  \frac{1}{2^{(i-1)(\alpha-1)}} \right)\right) + \frac{2^{\alpha+3} \kappa  F(\w_*)}{ T_{k+1}}  \\
= &   \frac{2^{\alpha^2+2\alpha+5} G^2}{\lambda (T_{k+1})^\alpha} +   \frac{2^{\alpha+3} \kappa  F(\w_*)}{ T_{k+1}} \left( \sum_{i=1}^{k+1}  \frac{1}{2^{(i-1)(\alpha-1)}} \right).
\end{split}
\]
\subsection{Proof of Theorem~\ref{thm:2}}
We first establish the following lemma for bounding the excess risk of the intermediate iterate.
\begin{lemma} \label{lemma:inter:new}  For any $k$, we have
\begin{equation} \label{eqn:risk:epoch:new}
\E[F(\w_1^{k+1})] -F(\w_*) \leq \frac{F(\w_1^1)-F(\w_*)}{2^{k}}+ \frac{F(\w_*)}{\beta} \left(\sum_{i=1}^{k} \frac{1}{2^{i-1}} \right).
\end{equation}
\end{lemma}

The number of epochs made is given by $k^\dag=  \lfloor  T/T' \rfloor$ and the final solution is $\wt=\w_1^{k^\dag+1}$. From Lemma~\ref{lemma:inter:new},  we have
\[
\begin{split}
&F(\w_1^{k^\dag+1})-F(\w_*)\\
 \leq  & \frac{F(\w_1^1)-F(\w_*)}{2^{k^\dag}}+ \frac{F(\w_*)}{\beta} \left(\sum_{i=1}^{k^\dag} \frac{1}{2^{i-1}} \right) \\
\leq &\frac{F(\w_1^1)-F(\w_*)}{2^{k^\dag}}+ \frac{2 F(\w_*)}{\beta} .
\end{split}
\]
\subsection{Proof of Lemma~\ref{lemma:inter:new}}
From (\ref{eqn:para:second}), we know that
\begin{eqnarray}
\eta L  = \frac{1}{4 \beta} \leq \frac{1}{4}, \label{eqn:eta:3}\\
\lambda \eta T'=  4. \label{eqn:eta:4}
\end{eqnarray}

We prove this lemma by induction on $k$. When $k=1$, from Lemma~\ref{lem:inner}, we have
\[
\begin{split}
&\E\left[ F(\w_1^{2})\right] - F(\w_*) \\
\leq &\frac{1}{2 \eta T' (1-2 \eta L)}  \|\w_{1}^1-\w_*\|^2 + \frac{2 \eta L }{(1-2 \eta L)} F(\w_*)\\
\overset{\text{(\ref{eqn:eta:3})}}{\leq} & \frac{1}{\eta T'}  \|\w_{1}^1-\w_*\|^2 + \frac{F(\w_*)}{\beta}\\
\overset{\text{(\ref{eqn:eta:4})}}{=} &\frac{\lambda}{4}   \|\w_{1}^1-\w_*\|^2 + \frac{F(\w_*)}{\beta}  \overset{\text{(\ref{eqn:stron:con})}}{\leq }  \frac{f(\w_{1}^1)-f(\w_*)}{2} + \frac{F(\w_*)}{\beta}.
\end{split}
\]

Assume that (\ref{eqn:risk:epoch:new}) is true for some $k \geq 1$, and we prove the inequality for $k+1$. According to Lemma~\ref{lem:inner}, we have
\[
\begin{split}
&\E\left[ F(\w_1^{k+2})\right] - F(\w_*) \\
\leq &\frac{1}{2 \eta T' (1-2 \eta L)}  \E \left[\|\w_{1}^{k+1}-\w_*\|^2\right] + \frac{2 \eta L }{(1-2 \eta L)} F(\w_*)\\
\overset{\text{(\ref{eqn:eta:3})}}{\leq} & \frac{1}{\eta T'}  \E \left[\|\w_{1}^{k+1}-\w_*\|^2\right] + \frac{F(\w_*)}{\beta} \\
\overset{\text{(\ref{eqn:eta:4})}}{=} &  \frac{\lambda}{4}  \E \left[\|\w_{1}^{k+1}-\w_*\|^2\right] + \frac{F(\w_*)}{\beta}  \\
\overset{\text{(\ref{eqn:stron:con})}}{\leq} &  \frac{\lambda}{4}  \frac{2}{\lambda}  \E \left[F(\w_{1}^{k+1})-F(\w_*)  \right]+ \frac{F(\w_*)}{\beta}  \\
\overset{\text{ (\ref{eqn:risk:epoch:new})}}{\leq} &  \frac{1}{2}   \left( \frac{F(\w_1^1)-F(\w_*)}{2^{k}}+ \frac{F(\w_*)}{\beta} \left(\sum_{i=1}^{k} \frac{1}{2^{i-1}} \right)\right)+ \frac{F(\w_*)}{\beta}  \\
=& \frac{F(\w_1^1)-F(\w_*)}{2^{k+1}}+ \frac{F(\w_*)}{\beta} \left(\sum_{i=1}^{k+1} \frac{1}{2^{i-1}} \right).
\end{split}
\]

\section{Conclusion and Future Work}
This paper aims to boost the convergence rate of stochastic approximation (SA) by exploiting smoothness and strong convexity  simultaneously. First, we prove an $O\left(1/[\lambda T^\alpha] +  \kappa  F_*/T\right)$ risk bound when $T =  \Omega(\kappa^\alpha)$. Thus, the convergence rate could approach $O(1/[\lambda T^\alpha])$ when the minimal risk is small. Second, we establish an  $O(1/2^{T/\kappa}+F_*)$ risk bound to further benefit from small risk. Thus, the excess risk reduces exponentially until reaching $O(F_*)$. We note that our proof is constructive and each risk bound is equipped with an efficient stochastic algorithm.

One limitation of this paper is that our risk bounds only hold in expectation. Although we can get a high-probability bound by introducing concentration inequalities \citep{Concentration_inequalities}, an $O(1/T)$ confidence term will appear in the upper bound, making it impossible to be faster than $O(1/T)$. To establish high-probability risk bounds, we may need more advanced mathematical tools or stronger assumptions, which will be investigated in the future.

\bibliography{E:/MyPaper/ref}

\appendix
\section{Comparison with \citet{NIPS2014_5355}}\label{sec:SGD}
First, we provide the following basic inequality that allows us to bound the excess risk by the distance. From Assumption~\ref{ass:1}, we have
\begin{equation}\label{eqn:apped:1}
F(\w_t)- F(\w_*) \leq \langle \nabla F(\w_*), \w_t -\w_*\rangle + \frac{L}{2} \| \w_t -\w_*\|^2.
\end{equation}

Using  notations of our paper, Theorem 2.1 of \citet{NIPS2014_5355} establishes the following convergence rate for unconstrained problems:
\begin{equation}\label{eqn:apped:2}
\E \left[ \|\w_t-\w_*\|^2 \right] \leq \left[ 1- 2 \gamma \lambda (1-\gamma L) \right]^T \|\w_0-\w_*\|^2 +  \frac{4 \gamma L F_*}{ \lambda (1-\gamma L) }
\end{equation}
where $\w_t$ is the  SGD iterate in the $t$-th round and $\gamma < 1/\lambda$ is the step size. Note that $\nabla F(\w_*)=0$ in the unconstrained case. Combining (\ref{eqn:apped:1}) and (\ref{eqn:apped:2}), we bound the expected risk as
\begin{equation}\label{eqn:apped:3}
\begin{split}
&\E\left[F(\wt)\right] - F(\w_*) \\
\overset{\text{(\ref{eqn:apped:1})}}{\leq}& \frac{L}{2} \E \left[ \|\w_t-\w_*\|^2 \right] \\
\overset{\text{(\ref{eqn:apped:2})}}{\leq}& \frac{L}{2} \left[ 1- 2 \gamma \lambda (1-\gamma L) \right]^T \|\w_0-\w_*\|^2 +  \frac{2\gamma L^2 F_*}{ \lambda (1-\gamma L) }.
\end{split}
\end{equation}
We have different ways to set the step size $\gamma$, and the convergence rate in (\ref{eqn:apped:3}) is always slower than ours.
\begin{compactitem}
\item By setting $\gamma=1/T$, we obtain an $O\big([1-\lambda/T]^T+ \kappa F_*/ T\big)$ rate, as shown in (\ref{eqn:rate:nips:2014}). This rate is worse than our $O\big(1/[\lambda T^2] + \kappa F_*/T\big)$ rate in Corollary~\ref{cor:1} because $[1-\lambda/T]^T$ becomes a constant when $T\rightarrow \infty$.
\item By setting $\gamma=1/(2L)$, the convergence rate is $O\big([1-1/\kappa]^T + \kappa F_*\big)$, as shown in (\ref{eqn:rate:nips:2014:second}). Although the first term decreases linearly, the second term has a  linear dependence on $\kappa$. So, it is slower than our $O(2^{-T/\kappa} + F_*)$ rate in Theorem~\ref{thm:2}.
\item When $F_*$ is known, we set
\[
\gamma= \frac{\epsilon \lambda }{ 2 \epsilon \lambda L + 8 L^2 F_*} \textrm{ and } T=  \Omega\left(\log \frac{1}{\epsilon} \cdot \frac{1}{\lambda \gamma} \right)=\Omega\left( \log \frac{1}{\epsilon} \left( \kappa + \frac{\kappa^2 F_*}{ \epsilon}\right) \right)
\]
to find an $\epsilon$-optimal solution. However, the above iteration complexity is higher than ours in (\ref{eqn:iter:our:fast}).
\end{compactitem}

For constrained problems, we can use projected SGD
\[
\w_{t+1}= \Pi_{\W}\left[\w_t-\gamma \nabla f_t(\w_t) \right]
\]
to enforce the domain constraint. Based on the nonexpanding property of the projection operator \citep{nemirovski-2008-robust}, it is easy to verify that (\ref{eqn:apped:2}) also hold when  projected SGD is used for constrained problems. Then, according to (\ref{eqn:apped:1}), we have the following upper bound for the expected risk
\begin{equation}\label{eqn:apped:4}
\begin{split}
&\E\left[F(\wt)\right] - F(\w_*)\\
\overset{\text{(\ref{eqn:apped:1})}}{\leq} & \|\nabla F(\w_*)\| \E\left[\|\w_t -\w_*\|\right] + \frac{L}{2} \E \left[ \|\w_t-\w_*\|^2 \right]\\
\leq &\|\nabla F(\w_*)\| \sqrt{\E \left[ \|\w_t-\w_*\|^2 \right]} + \frac{L}{2} \E \left[ \|\w_t-\w_*\|^2 \right]
\end{split}
\end{equation}
where the last step is due to Jensen's inequality \citep{Convex-Optimization}. Then, we can bound the expected risk by substituting (\ref{eqn:apped:2}) into (\ref{eqn:apped:4}). However, because of the square root operation, the convergence rate is slower than that in (\ref{eqn:apped:3}) of the unconstrained case, and thus slower than our rate which holds for both constrained and unconstrained problems.

\end{document}